\documentclass[11pt]{article}
\usepackage{coling2020}
\usepackage{times}
\usepackage{url}
\usepackage{latexsym}

\usepackage{graphicx}
\usepackage{subfig}
\usepackage{amsmath}
\usepackage{amssymb}
\usepackage{multirow,paralist}
\usepackage{bm}

\usepackage{color,xcolor}

\colingfinalcopy 

\title{Adversarial Learning on the Latent Space for Diverse Dialog Generation}

\author{Kashif Khan\textsuperscript{1}\thanks{\ \ Equal contribution.}, Gaurav Sahu\textsuperscript{1}\footnotemark[1], Vikash Balasubramanian\textsuperscript{1}\footnotemark[1], Lili Mou\textsuperscript{2}, Olga Vechtomova\textsuperscript{1} \\
\textsuperscript{1}University of Waterloo\\ \textsuperscript{2}Dept.~Computing Science, University of Alberta\\ Alberta Machine Intelligence Institute (Amii)  \\
\tt{\{kashif.khan,gaurav.sahu,v9balasu,ovechtom\}@uwaterloo.ca} \\
\tt{doublepower.mou@gmail.com} \\
}

\date{}

\begin{document}
\maketitle
\begin{abstract}
    Generating relevant responses in a dialog is challenging, and requires not only proper modeling of context in the conversation,
    but also being able to generate fluent sentences during inference.
    In this paper, we propose a two-step framework based on generative adversarial nets for generating conditioned responses.
    Our model first learns a meaningful representation of sentences by autoencoding, and then learns to map an input query to the response representation, which is in turn decoded as a response sentence.
    Both quantitative and qualitative evaluations show that our model generates more fluent, relevant, and diverse responses than existing state-of-the-art methods.\footnote{The code is available at \url{https://github.com/vikigenius/conditional_text_generation}}\blfootnote{
    \hspace{-0.65cm}  
     This work is licensed under a Creative Commons 
     Attribution 4.0 International License.
     License details:
     \url{http://creativecommons.org/licenses/by/4.0/}.
}
\end{abstract}

\section{Introduction}
\label{ref:intro}

Dialog generation is a challenging problem because it not only requires us to model the context in a conversation but also to exploit it to generate a relevant and fluent response.
    A dialog generation system can be divided into two parts:
    1) encoding the context of the conversation,
    and 2) generating a response conditioned on the given context.
    A generated response is considered to be ``good" if it is meaningful, fluent, and most importantly, relevant to the given context. 

    With the advancement of deep learning, sequence-to-sequence (Seq2Seq) models \cite{sutskever2014sequence} are adopted for dialog systems to encode conversational context and  generate a response.
    However, they suffer from the problem of generic utterance generation, e.g., always generating ``I don't know'' \cite{serban2015hred,li-etal-2016-diversity}. 
    One possible explanation~\cite{why} is the high uncertainty in dialog generation.
    A plausible response is analogous to a ``mode'' of a continuous distribution, and the response distribution is thus multimodal.
    However, the decoder of a  Seq2Seq model is trained by cross-entropy loss, which is equivalent to minimizing the KL divergence between the target and predicted distributions.
    The asymmetric nature of KL divergence makes the learned distribution wide-spreading, analogous to the mode-averaging problem for continuous variables.
    
    Variational encoder-decoders \cite{serban2016vhred,bahuleyan2017variational,zhao2017cvae} and Wasserstein encoder-decoders \cite{bahuleyan2018stochastic} adopt probabilistic modeling to encourage diversity in responses.
    However, their decoders are also trained by cross-entropy loss against the target sequence, still making the model generate generic utterances.

    In this paper, we propose an approach that uses adversarial learning in the latent space for dialog generation.
    We first train a variational autoencoder (VAE) \cite{kingma2013auto} on sentences, and then apply a generative aderversarial network (GAN) on the latent space of the VAE.
    At inference time, we obtain the latent representation of the response from the generator of the GAN and decode it using the VAE's decoder. In this way, we can benefit from the mode-capturing property of GANs \cite{mao2019mode,thanh2019improving}.
    Also, our GAN is trained on the latent space, and techniques like Gumbel-Softmax and reinforcement-learning (RL) are not necessary, which
    largely simplifies the training procedure.
    We further introduce a mean squared error (MSE) auxiliary loss to our adversarial module, which mitigates the mode-missing problem in GANs \cite{che2016mode}, resulting in more relevant and diverse responses.

    We evaluate our model on the deduplicated version \cite{bahuleyan2017variational} of the benchmark DailyDialog dataset \cite{li-etal-2017-dailydialog} and also the Switchboard dataset \cite{Godfrey1997SwitchBoard}. 
    Results indicate that responses generated by our model are more relevant to the input query/context, and are more diverse and fluent than the existing baselines.
    
   The main contributions of our paper are as follows.
   \begin{enumerate}
       \item We propose a two-step framework of latent-space adversarial learning for generating diverse and relevant responses.
       \item We propose a combination of adversarial loss and an auxiliary mean squared loss to help the GAN to converge faster and achieve better performance for dialog generation.
   \end{enumerate}

    \begin{figure*}[t]
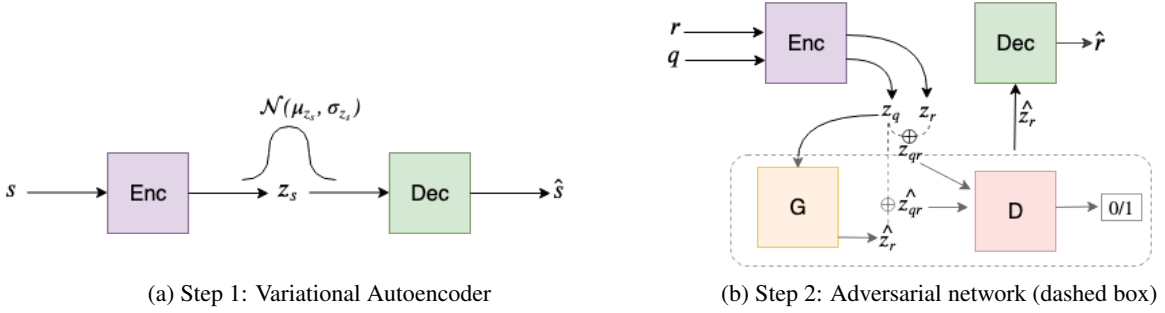

        \centering
        \subfloat[Step 1: Variational Autoencoder ]{\includegraphics[scale=0.5, trim={-7mm -10mm 6mm 0}]{figures/Step1.png}}
        \subfloat[Step 2: Adversarial network (dashed box)]{\includegraphics[scale=0.5]{figures/Step2.png}}
        \caption{The framework of our proposed two-step training procedure.   $\bigoplus$ denotes concatenation.}
        \label{fig:two_step}
    \end{figure*}

\section{Approach}
\label{sec:appr}

Figure \ref{fig:two_step} provides an overview of our proposed two-step approach. 

\textbf{Step 1:} We first train an autoencoder, which takes an utterance $s$ (either a query or a response) as input, gets its latent code $\bm z_s$ from the encoder, and then feeds it to a decoder for reconstructing. The autoencocder learns a real-valued vector representation of a generic sentence.

\textbf{Step 2:} We train an adversarial network on the latent $\bm z$ space for learning dialog generation. Given a query-response pair $(q, r)$ in the training set, we use the trained encoder from Step 1 to obtain their latent variables $\bm z_q$ and $\bm z_r$. The query latent variable $\bm z_q$ is fed to a generator $G$, that maps it to the corresponding response's latent variable $\hat{\bm z}_r$. When training the generator, we aim to match $\bm z_r$ and $\hat{\bm z}_r$ through the adversarial loss combined with a mean-squared error loss. In here, the adversarial loss further involves a discriminator that classifies the predicted response representation $\hat{\bm z}_r$ versus the encoded representation of the true response $\bm z_r$, conditioned on the query $\bm z_q$. 

The details of our approach will be introduced in the rest of this section.

    \subsection{Step 1: Training an Autoencoder} \label{step_1}
    
    In Step 1, our primary goal is to learn a continuous representation of all utterances in the dialog corpus. The mapping from a sentence to its continuous representation should ideally be invertible so that our adversarial loss (in Step 2) could be applied to the continuous space to generate dialog responses.
    
    In particular, we adopt a variational autoencode~\cite[VAE]{kingma2013auto} for our first step. A VAE encodes an input sentence $s$ to a probabilistic, latent continuous representation $\bm z$, from which the input sentence $s$ is reconstructed. 
    
    We first impose a prior distribution on $\bm z$, which is typically set to standard normal $p(\bm z)=\mathcal{N}(\bm 0, \mathbf I)$. Given the sentence $s$, VAE encodes a posterior distribution $q_E(\bm z|s)=\mathcal{N}(\bm\mu, \operatorname{diag}\bm \sigma^2)$, where $\bm\mu$ and $\bm\sigma$ are predicted by the encoder of VAE. The training objective is to minimize the expected reconstruction loss, penalized by a KL divergence term between the posterior and the prior. This is given by
    \begin{equation}
        J_{\text{AE}}(\bm \theta_\text{Enc}, \bm \theta_\text{Dec}) = - \mathbb{E}_{q_E(\bm z|s)}[\textrm{log}\ p(s|\bm z)]
        +\lambda_\text{KL} \textrm{KL}(q_E(\bm z|s)||p(\bm z))
    \label{eq:our_vae_loss}
    \end{equation}
    where $\lambda_\text{KL}$ balances the two terms. 
    
    Compared with a deterministic autoencoder, VAE learns a smoother latent space by its KL regularization. This is helpful during the second step, where a GAN is trained to predict the latent representation of a response for decoding.

    \subsection{Step 2: Predicting the Representation of the Response} \label{step_2}
    
    In Step 2, the main objective is to predict the representation of the response given dialog context (such as the previous utterance). In this way, the predicted latent representation of the response can be fed to the trained decoder in Step 1 to generate the response utterance. 
    
    To predict the response representation, we re-use the encoder in Section~\ref{step_1} to capture the meaning of the context query as $\bm z_q$.
    Then we have a two-layer perceptron (with a ReLU activation function in the hidden layer) to predict the representation of the utterance to be generated, denoted by $\hat{\bm z}_r={G}(\bm z_q)$. 
    
    For adversarial training, we also encode the representation of the ground truth reply $r$ as $\bm z_r$ by the encoder in Step 1.
    We train an adversarial discriminator $D$ to classify whether the response representation is real or predicted. Such classification should be based on context, because we would learn not only if an utterance is appropriate as a reply, but also if the utterance is appropriate to a specific query. Therefore, we also feed the encoded context representation into the discriminator. The classification is denoted by $D(\bm z_r,\bm z_q)$ or $D(\hat{\bm z}_r,\bm z_q)$, where we essentially concatenate the representations of the response and the query before feeding them to a logistic regression layer. 
    
    The adversarial loss for training the latent space is given by:
    \begin{align}
        J_{\text{CGAN}} &= \min_G \max_D V(D, G)\\
        V(D, G) &= \mathbb{E}_{ (\bm z_q,\bm z_r)\sim\mathcal{D}_\text{train}}[\textrm{log} D(\bm z_r, \bm z_q)
        +\textrm{log}(1 - D(G(\bm z_q), \bm z_q))]
    \end{align}
    where $\mathcal{D}_\text{train}$ is the training data.
    
    In other words,  the discriminator $D$ is trained by maximizing $V(D, G)$ so as to distinguish the true representation of a response and the predicted response representation given the query, whereas the generator $G$ is trained to fool the discriminator by minimizing $V(D, G)$.
    
    It should be emphasized that our model is different from adversarial autoencoders \cite{DBLP:journals/corr/MakhzaniSJG15}, because our discriminator takes the query into consideration. Our adversarial loss learns an implicit conditional distribution $p(\bm z_r|\bm z_q)$, instead of a marginal distribution $p(\bm z_r)$ as in \newcite{pmlr-v80-zhao18b}.
    
    Additionally, we introduce an auxiliary mean square error (MSE) loss to the objective function:
    \begin{equation}
        J_{\text{MSE}}  = ||\bm z_r -  \hat{\bm z}_r||^2
    \end{equation}
     
    The MSE loss on the generator helps stabilize the GAN training and mitigate the mode-missing problem of GANs \cite{che2016mode}. In summary, the overall training objective is given by
    \begin{equation}
        J = J_{\text{CGAN}} + \gamma J_{\text{MSE}} 
        \label{eq:our_dm_loss}
    \end{equation}
    where $\gamma$ is a tunable hyperparameter that moderates the effect of the MSE loss.
    
    For inference, our model first uses the pretrained VAE from Step 1 to encode an unseen query $q^*$ as $\bm z_{q^*}$.
    This encoded representation is then passed to the generator $G$ to predict the response latent code $G(z_{q^*})$, which is finally fed to the decoder of the VAE from Step 1 to generate a response sentence.
    
   In our experiments, we have two settings for dialog generation: single-turn and multi-turn. In the single-turn setting, we form query-response samples by extracting every pair of consecutive utterances of a conversation in the training data.
    
 In the multi-turn setting, we have the query-response pairs by extracting every utterance with its preceding utterances in the \textit{entire} conversation.
The VAE in Step 1 remains the same, but we introduce another RNN to encode context. Specifically, it is built upon the VAE's encoded representation of each utterance, and yields a fixed-length vector representation of the entire context. 
During the adversarial training, 
we concatenate the context vector with the query (immediate preceding utterance) representation before feeding them to the generator.
In this way, our generator now also takes the context into account when predicting the response latent code.
A similar adjustment is applied during inference as well.
    
    \begin{table*}[t]
        \small
        \centering
        \begin{tabular}{c|c|c|c|c|c|c|c|c|c|c}
        \hline
            \multirow{2}{*}{\textbf{Model}} 
            & \multicolumn{3}{|c|}{\textbf{BLEU}} & \multicolumn{6}{c|}{\textbf{Diversity}} & \textbf{Fluency} \\
            \cline{2-11}
            & \textbf{Avg} & \textbf{Max} & \textbf{HM} & \textbf{Intra-1} & \textbf{Intra-2} & \textbf{Inter-1} & \textbf{Inter-2} & \textbf{ASL (14.43)} & \textbf{TTR} & \textbf{PPL}\\
            \hline
     \multicolumn{11}{c}{\textbf{Single-turn results}} \\ \hline

            \textbf{Seq2Seq} & 0.143 & 0.217	& 0.172	&	\textbf{0.99} & \textbf{0.99} & 0.46 & 0.49 & 4.63 & 0.019 & 18.45 \\ \hline
            \textbf{WED-S} & 
            0.215 & 0.357 & 0.268 & 0.94 & \textbf{0.99} & 0.48 & 0.74 & 10.42 & 0.034 & 33.91 \\
            \hline
            \textbf{DialogWAE} & 0.296	& 0.356 & 0.323 &	0.85 & 0.97 & 0.42 & 0.74 & 19.34 & 0.005 & 20 \\ \hline
            \textbf{VAE-M (ours)} & 0.191 & 0.293 & 0.231	&	0.98 & \textbf{0.99} & \textbf{0.5} & 0.79 & 9.36 & 0.029 & 19.7      \\ \hline
            \textbf{VAE-A (ours)} & 0.295 & 0.359	& 0.323	&	0.93 & \textbf{0.99} & 0.46 & 0.76 & 13.64 & 0.035 & 21.38 \\ \hline
            \textbf{VAE-AM (ours)} & \textbf{0.306} & \textbf{0.367} & \textbf{0.334}	&	0.91 & \textbf{0.99} & 0.46 & \textbf{0.82} & 16.90 & \textbf{0.034} & \textbf{17.01}
            \\ \hline
            \multicolumn{11}{c}{\textbf{Multi-turn results}} \\ \hline
                \textbf{HRED*} & 0.232	& 0.232 & 0.232 & 0.94 & 0.97 & 0.09 & 0.10 & 10.1 & - & - \\  
                \hline
                \textbf{CVAE*}	& 0.222 & 0.265 & 0.242 & 0.94 & 0.97 & 0.18 & 0.22 & 10.0 & - & - \\
                \hline
                \textbf{CVAE-CO*} & 0.244  & 0.259 & 0.251 & 0.82 & 0.91 & 0.11 & 0.13 & 11.2 & - & - \\
                \hline
                \textbf{VHCR*} & 0.266	& 0.289 & 0.277 & 0.85 & 0.97 & 0.42 & 0.74 & 16.9 & - & - \\
                \hline
                \textbf{DialogWAE} & 0.279	& 0.365 & 0.316 &	0.79 & 0.92 & 0.35 & 0.68 & 19.84 & 0.007 & 161.86 \\
                \hline
                \textbf{VAE-AM (ours)} & \textbf{0.314} & \textbf{0.371} & \textbf{0.340}	&	\textbf{0.847} & \textbf{0.98} & \textbf{0.41} & \textbf{0.73} & 15.3 & \textbf{0.036} & \textbf{119.39}   \\ \hline
        \end{tabular}
        \caption{Results on the DailyDialog dataset. BLEU scores are computed by the average/maximum of 10 randomly sampled replies. HM is the harmonic mean of average and maximum BLEU scores. Suffix A: adversarial loss; suffix M: MSE loss; suffix AM: both adversarial and MSE losses. * denotes results taken from \newcite{gu2018dialogwae}, whose training and test sets contain duplicate samples. The Bold font shows the best performance on the de-duplicated dataset. The number in the bracket of the ASL column is the groundtruth average sentence length.
        }
        \label{tab:dialog_results}
    \end{table*}
    
\begin{table*}[t]
    \small
    \centering
    \begin{tabular}{c|c|c|c|c|c|c|c|c|c|c}
    \hline
        \multirow{2}{*}{\textbf{Model}} 
        & \multicolumn{3}{|c|}{\textbf{BLEU}} & \multicolumn{6}{c|}{\textbf{Diversity}} & \textbf{Fluency} \\
        \cline{2-11}
        & \textbf{Avg} & \textbf{Max} & \textbf{HM} & \textbf{Intra-1} & \textbf{Intra-2} & \textbf{Inter-1} & \textbf{Inter-2} & \textbf{ASL (8.49)} & \textbf{TTR} & \textbf{PPL}\\
        \hline
        \multicolumn{11}{c}{\textbf{Single-turn results}}  \\ \hline
        \textbf{Seq2Seq} & 0.088 & 0.176 & 0.118 & \textbf{0.989} & 0.956 & \textbf{0.816} & \textbf{0.927} & 2.66 & 0.026 & 23.62
        \\
        \hline
        \textbf{WED-S} & 0.193 & \textbf{0.395} & 0.259 &	0.941 & 0.989 & 0.404 & 0.525 & 10.41 & 0.032 & 35.63   \\ \hline
         \textbf{DialogWAE} & 0.235  & 0.375 & 0.289 & 0.739 & 0.712 & 0.354 & 0.571 & 10.32 & 0.017 & 25.36 \\
        \hline
        \textbf{VAE-M (ours)} & 0.231 & 0.3 & 0.261 & 0.954 & 0.998 & 0.322 & 0.479 & 8.84 & 0.045 & 24.58 \\ \hline
        \textbf{VAE-A (ours)} & 0.229 & 0.376 & 0.285 & 0.725 & 0.751 & 0.218 & 0.354 & 11.73 & 0.053 & 27.55 \\
        \hline
        \textbf{VAE-AM (ours)} & \textbf{0.259} & 0.364 & \textbf{0.303} & \textbf{0.989} & \textbf{0.999} & 0.436 & 0.569 & 7.08 & \textbf{0.062} & \textbf{21.87} \\
        \hline
        \multicolumn{11}{c}{\textbf{Multi-turn results}} \\ \hline
            \textbf{HRED*} & 0.262	& 0.262 & 0.262 & 0.813 & 0.452 & 0.081 & 0.045 & 12.1 & - & - \\  
            \hline
            \textbf{CVAE*}	& 0.258 & 0.295 & 0.275 & 0.803 & 0.415 & 0.112 & 0.102 & 12.4 & - & - \\
            \hline
            \textbf{CVAE-CO*} & 0.269  & 0.299 & 0.283 & 0.863 & 0.581 & 0.111 & 0.110 & 10.3 & - & - \\
            \hline
            \textbf{VHCR*} & 0.234 & 0.276 & 0.254 & 0.877 & 0.536 & 0.130 & 0.131 & 9.29 & - & - \\
            \hline
            \textbf{DialogWAE} & 0.267	& \textbf{0.394} & \textbf{0.318} & 0.779 & 0.844 & 0.325 & 0.513 & 14.7 & 0.019 & 183.82 \\
            \hline
            \textbf{VAE-AM (ours)} & \textbf{0.271} & 0.372 & 0.313	& \textbf{0.954} & \textbf{0.966} & \textbf{0.412} & \textbf{0.559} & 8.72 & \textbf{0.065} & \textbf{136.81} \\ \hline
    \end{tabular}
    \caption{Results on the Switchboard dataset. * denotes the numbers taken from \newcite{gu2018dialogwae}.}
    \label{tab:swda_results}
\end{table*}

\section{Experiments}
\label{sec:exp}

We conduct experiments on the DailyDialog dataset \cite{li-etal-2017-dailydialog}, a manually labeled multi-turn dialog dataset, and the Switchboard dataset \cite{Godfrey1997SwitchBoard}, a dialog dataset containing transcripts of telephonic conversations. For DailyDialog, we use the original splits after removing duplicates following \newcite{bahuleyan2018stochastic}.
We use the the AllenNLP framework \cite{Gardner2017AllenNLP} to implement all our models. Appendix~\ref{app:params} presents more experimental details and hyper-parameters.

We use the following baseline models for comparison:
\begin{itemize}
    \item \textbf{Seq2Seq.} The standard sequence to sequence model based on LSTM.
    \item \textbf{WED-S.} A stochastic Wasserstein encoder-decoder model  \cite{bahuleyan2018stochastic}.
    \item \textbf{DialogWAE.} A model based on adversarial regularization of autoencoders \cite{gu2018dialogwae}.
    \item \textbf{HRED.} A generalized Seq2Seq model that uses hierarchical RNN encoder \cite{serban2015hred}.
    \item \textbf{CVAE.} A conditional VAE model with KL annealing \cite{zhao2017cvae}.
    \item \textbf{CVAE-CO.} A collaborative conditional VAE model  \cite{shen2018cvaeco}.
\end{itemize}

    \subsection{Results and Analysis}
    
    The results for the Daily Dialog and the Switchboard datasets are shown in Tables \ref{tab:dialog_results} and \ref{tab:swda_results}, respectively.
    The generated responses are evaluated by the following criteria:
    
    \textbf{Overall quality.} We measure the quality of the generated responses by BLEU scores \cite{Papineni2002bleu}, for which we adopt the smoothing techniques in \newcite{gu2018dialogwae}. For each query, we generate 10 responses for a query, and compute the average and maximum BLEU scores.
    Then we also compute the harmonic mean of the average and the maximum BLEU scores.\footnote{The evaluation protocol follows \newcite{gu2018dialogwae} with code at \url{https://github.com/guxd/DialogWAE/blob/29f206af05bfe5fe28fec4448e208310a7c9258d/experiments/metrics.py#L90}}
    Our model is either the best-performing model or highly competitive in terms of the BLEU scores.
    The DialogWAE model also achieves high BLEU scores, while the Seq2Seq model is the worst-performing model.
    
    \textbf{Diversity.} We measure the diversity of dialog generation in two aspects:
    \begin{compactitem}
        \item \textbf{Intra-diversity}. The Intra-diversity score measures the proportion of distinct unigrams and bigrams in each response. It is similar for most models.
        \item \textbf{Inter-diversity}. The Inter-diversity scores measure the proportion of distinct unigrams and bigrams across all 10 responses. 
    \end{compactitem}
    We note that our model performs the best across Inter-diversity metrics. 
    We further use other diversity indicators such as the Average Sentence Length (ASL) of the responses.
    We see that diversity scores for the Seq2Seq model are very high on the Switchboard dataset; however, it has the lowest ASL score as well.
    This observation is within expectations, and the Seq2Seq model does not generate diverse responses overall.
    DialogWAE generates longer responses on average; however, our model is closer to the ground truth ASL (14.43 for DailyDialog and 8.49 for Switchboard). We also note that our model achieves good Type-Token Ratio (TTR) scores,\footnote{TTR is computed in the corpus level, whereas Inter-$n$ diversity is the average of per-sample distinct unigram ratio.} indicating diverse word choices when compared with other models.

    \textbf{Fluency.} We compute the PPL scores of generated responses to measure fluency.
    We notice that our model achieves the best PPL scores, although DialogWAE is quite close.
    The Seq2Seq model also achieves low PPL, but this is mainly due to the short and generic responses.
    Interestingly, PPL scores are generally higher in the multi-turn setting, which may be attributed to the increased complexity of the output when more context is given.
    
    \textbf{Analysis of Losses.} Combining the MSE and adversarial losses leads to significant improvements across all metrics, including the BLEU scores, response diversity (Inter-1 and Inter-2), and fluency (PPL).
    In our experiments, we also notice that the MSE term leads to quicker and more stable convergence of the GAN (within 6 epochs), making training easier.

We present human evaluation in Appendix~\ref{app:human} and a case study in Appendix~\ref{app:case}.
\section{Conclusion}
\label{sec:cncl}

We propose an effective two-stage model for dialog generation. We make use of sentence representations learned by a VAE and train a adversarial network on VAE's latent space to generate diverse responses given a query and context.
We observe that our model outperforms existing state-of-the-art approaches by generating more diverse, fluent, and relevant sentences.

\section*{Acknowledgments}
This research was partially supported by the Natural Sciences and Engineering Research Council of Canada
(NSERC) under grant Nos.~RGPIN-2019-04897
and RGPIN-2020-04465. Lili Mou is also supported by the Amii Fellow Program and
the Canadian CIFAR AI Chair Program. This research was enabled in part by the support of Compute Canada (www.computecanada.ca).

\bibliographystyle{coling}
\bibliography{coling2020}

\newpage
\appendix


\section{Hyperparameter Settings and Training}
\label{app:params}
\textbf{Single-turn.} In this setting, we first train a VAE on the entire corpus.
    We use a single-layer encoder with Bidirectional LSTMs \cite{hochreiter1997long} and a unidirectional LSTM layer for the decoder of the VAE.
    Both use a hidden size of 512.
    The dimension of our latent vectors is 128, and that of the word embeddings is 300.
    Further, we adopt KL-annealing and word dropout from \newcite{bowman-etal-2016-generating} to stabilize VAE's training.
    We use a word dropout probability of 0.5 and a sigmoid annealing schedule to anneal the KL weight to 0.15 for 4500 iterations. The performance statistics of VAE in Step 1 are shown in \ref{tab:vae_stats}.

\begin{table}[htb]\centering
\begin{tabular}{|c|c|c|c|c|}
\hline
Model & KL & BLEU & Dist-1 & Dist-2\\
\hline
VAE & 18.8 & 0.18 & 0.32 & 0.49\\
\hline
\end{tabular}
\caption{Performance of the VAE in Step 1 on the DailyDialog dataset. BLEU is the reconstruction BLEU-4, Dist-1 and Dist-2 are distinct unigrams and bigrams in generated samples, KL is the validation KL in the best epoch (ELBO).}
\label{tab:vae_stats}
\end{table}

    For the GAN, we use a 2-layer feed-forward network  with a hidden layer of 256 units as the generator, along with batch normalization \cite{ioffe2015batch} and LeakyReLU activation \cite{Maas13}.
    The discriminator shares a similar architecture.
    We use Adam \cite{kingma2014adam} to optimize all our networks.

\textbf{Multi-turn.} In this setting, the VAE's architecture remains the same as the Single-turn setting.
We introduce another BiLSTM encoder with hidden size of 512, which is fed with the VAE-encoded representations of the context sentences.
Other hyperparameters are kept the same.
For implementation, our generator predicts the response representation at each turn, but we use teacher-forcing, assuming the context is the actual previous utterances.
\section{Human Evaluation}
\label{app:human}
In addition to automatic metrics, we also evaluate our model and compare it with DialogWAE \cite{gu2018dialogwae} using human evaluation.
Five human judges rate the response of each model on a scale of 1--5, according to two criteria: 1) Relevance to the query, and 2) Fluency of the generated response.
As can be seen from Table \ref{tab:human} our model is competitive and achieves better scores on both criteria.

\begin{table}[ht]
    \centering
    \begin{tabular}{|c|c|c|}
    \hline
        Model & Relevance & Fluency  \\
        \hline
        VAE-AM & 2.3625 & 3.2225 \\
        DialogWAE & 2.3125 & 3.1775 \\
        \hline
    \end{tabular}
    \caption{Human evaluation results }
    \label{tab:human}
\end{table}

\section{Case Study}
\label{app:case}
We conduct a case study on VAE-AM's responses with DialogWAE's in the single-turn and multi-turn settings.
For both settings, we show the best response out of 10 generated responses for each input.

Table \ref{tab:gen_single} shows that DialogWAE generates generic responses, although its sentences are longer. Our model has more concrete concepts, such as ``leaflets'' and ``money.''

Table 6 shows generated responses in the multi-turn setting.
We see that both models are more informative as we have more context compared with Table \ref{tab:gen_single}.
However, DialogWAE is repeating the same concept, such as ``a lot of'' in Examples 2, 3, and 5, and ``have to pay" in Examples 2 and 5. 
This observation is also consistent with the TTR results in Tables \ref{tab:dialog_results} and \ref{tab:swda_results}.

\begin{table}[t]
    \small
    \centering
    \begin{tabular}{| l | l |}
    \hline
        \textbf{\#} & \textbf{Generated Responses} \\
        \hline
        \multirow{3}{*}{\textbf{1}} & \textbf{Query:} o . k . well , go to google.com , and type in battle of gettysburg. \\
        & \textbf{VAE-AM:} oh , there are a lot of things that i need to do . \\
        & \textbf{DialogWAE:} then ill have to check the details . what can i do for you today ? \\
        \hline
        \multirow{3}{*}{\textbf{2}} & \textbf{Query:} yes . i m going to the untied states for travel . what places do you recommend ? \\
        & \textbf{VAE-AM:} here are some leaflets detailing everything . i have nt seen you in ages . \\
        & \textbf{DialogWAE:} it s very nice to see you again , but i think it s a good idea to do . \\
        \hline
        \multirow{3}{*}{\textbf{3}} & \textbf{Query:} medium one will be perfect , thank you \\
        & \textbf{VAE-AM:} is that all ? \\
        & \textbf{DialogWAE:} in that case , i have to be able to get some of them . \\
        \hline
        \multirow{3}{*}{\textbf{4}} & \textbf{Query:} the annual charge rate is NUM . \\
        & \textbf{VAE-AM:} that s a lot of money . can you make it a little cheaper ? \\
        & \textbf{DialogWAE:} then what about the \texttt{<UNK>} ? i mean , i am not sure . \\
        \hline
        \multirow{3}{*}{\textbf{5}} & \textbf{Query:} but it is too loose.do you have the smaller size ? \\
        & \textbf{VAE-AM:} ok , we will . how about this one ? \\
        & \textbf{DialogWAE:} in that case , i have to be able to get some of them . i think i can get a new one . \\
        \hline
    \end{tabular}
    \caption{Generated responses in the single-turn setting}
    \label{tab:gen_single}

\end{table}

\begin{table}[t]
    \small
    \centering
    \begin{tabular}{| l | l |}
    \hline
        \textbf{\#} & \textbf{Generated Responses} \\
        \hline
        \multirow{3}{*}{\textbf{1}} & \textbf{Query:} allan, i am really disappointed by your lack of effort on this project . \\
        &
        \textbf{VAE-AM:} it seems you are not aware of the actual work split . i have done more than anyone else. \\
        &
        \textbf{DialogWAE:} ok , it's not that bad . but it's not that bad . i don't want to have any other questions about it . \\
        \hline
        \multirow{3}{*}{\textbf{2}} & \textbf{Query:} when is he coming ? \\
        &
        \textbf{VAE-AM:} next week . he is enjoying his month long vacation in miami . \\
        &
        \textbf{DialogWAE:} sorry , he's not very serious . he's got a lot of money to do .  \\
        \hline
        \multirow{3}{*}{\textbf{3}} & \textbf{Query:} i ll try my best to make a good impression on the interviewer . \\
        &
        \textbf{VAE-AM:} do nt worry pal. only weak candidates should. just dress properly and try to make it alive there haha ... \\
        &
        \textbf{DialogWAE:} that is ok , we ’ ll be able to get a raise in the next few days . we will have to pay you a lot of money . \\
        \hline
        \multirow{3}{*}{\textbf{4}} & \textbf{Query:} can you tell me where the pots and pans are ? \\
        &
        \textbf{VAE-AM:} absolutely not . i m not letting you cook today . you messed it real bad yesterday . \\
        &
        \textbf{DialogWAE:} yes , of course .  \\
        \hline
        \multirow{3}{*}{\textbf{5}} & \textbf{Query:} what kind of personal information ? \\
        &
        \textbf{VAE-AM:} your name , height , sex ,  and contact . also tick this check box if you have faced this issue before . \\
        &
        \textbf{DialogWAE:} not at the moment . but there is a lot of money in the city , but we have to pay a lot of money for the money . \\
        \hline
    \end{tabular}
    \caption{Generated responses in the multi-turn setting}
    \label{tab:gen_multi}
\end{table}

\end{document}


\appendix

\section{Appendix}
In this section, we describe the training details and hyperparameters employed to train the single-turn and multi-turn VAE-AM network.
Appendix \ref{app:gen} compares the outputs of our model with DialogWAE for the two settings.

\subsection{Parameter Settings and Training}
\textbf{Single-turn.} In this setting, we first train a VAE on the entire corpus.
    We use a single-layer encoder with Bidirectional LSTMs \cite{hochreiter1997long} and a unidirectional LSTM layer for the decoder of the VAE.
    Both use a hidden size of 512.
    The dimension of our latent vectors is 128, and that of the word embeddings is 300.
    Further, we adopt KL-annealing and word dropout from \newcite{bowman-etal-2016-generating} to stabilize VAE's training.
    We use a word dropout probability of 0.5 and a sigmoid annealing schedule to anneal the KL weight to 0.15 for 4500 iterations. The performance statistics of stage 1 VAE are shown in \ref{tab:vae_stats}.

\begin{table}[htb]\centering
\begin{tabular}{|c|c|c|c|c|}
\hline
Model & KL & BLEU & Dist-1 & Dist-2\\
\hline
Vae & 18.8 & 0.18 & 0.32 & 0.49\\
\hline
\end{tabular}
\caption{Performance stats of stage 1 VAE. BLEU is the reconstruction BLEU-4, Dist-1 and Dist-2 are distinct Unigrams and Bigrams in generated samples, KL is the validation KL in the best epoch (ELBO).}
\label{tab:vae_stats}
\end{table}

    For the GAN, we use a 2-layer feed-forward network  with a hidden layer of 256 units as the generator, along with batch normalization \cite{ioffe2015batch} and LeakyReLU activation \cite{Maas13}.
    The discriminator shares a similar architecture.
    We follow standard GAN tricks from \newcite{goodfellow2016nips} and train our GAN till convergence using teacher forcing.
    We use Adam \cite{kingma2014adam} to optimize all our networks.

\textbf{Multi-turn.} In this setting, the VAE's architecture remains the same as the Single-turn setting.
However, we introduce a BiLSTM layer with hidden size of 512 to the generator to capture the context.
Rest of the hyperparameters are kept the same.
It is important to note that our network generates a latent vector at each turn, but since we are using teacher-forcing, the final output is ultimately conditioned on the actual context.

\subsection{Generated Samples}
\label{app:gen}
We compare VAE-AM's responses with DialogWAE's in the single turn (Table \ref{tab:gen_single}) and multi-turn setting (Table \ref{tab:gen_multi}).
For both the settings, we show the best response out of the 10 generated responses.
We observe that our model generates more informative responses, in general.

\begin{table*}[t]
    \small
    \centering
    \begin{tabular}{c}
    \hline
        \textbf{Generated Responses} \\
        \hline
        \textbf{Query:} o . k . well , go to google.com , and type in battle of gettysburg . \\
        \hline
        \textbf{VAE-AM:} oh , there are a lot of things that i need to do . \\
        \textbf{DialogWAE:} then ill have to check the details . what can i do for you today ?  \\
        \hline
        \textbf{Query:} yes . i m going to the untied states for travel . what places do you recommend ? \\
        \hline
        \textbf{VAE-AM:} here are some leaflets detailing everything . i have nt seen you in ages . \\
        \textbf{DialogWAE:} it s very nice to see you again , but i think it s a good idea to do . \\
        \hline
        \textbf{Query:} medium one will be perfect , thank you \\
        \hline
        \textbf{VAE-AM:} is that all ? \\
        \textbf{DialogWAE:} in that case , i have to be able to get some of them . \\
        \hline
        \textbf{Query:} the annual charge rate is NUM . \\
        \hline
        \textbf{VAE-AM:} that s a lot of money . can you make it a little cheaper ? \\
        \textbf{DialogWAE:} then what about the \texttt{<UNK>} ? i mean , i am not sure . \\
        \hline
        \textbf{Query:} but it is too loose.do you have the smaller size ? \\
        \hline
        \textbf{VAE-AM:} ok , we will . how about this one ? \\
        \textbf{DialogWAE:} in that case , i have to be able to get some of them . i think i can get a new one . \\
        \hline
    \end{tabular}
    \caption{Generated responses in single-turn setting}
    \label{tab:gen_single}
\end{table*}

\begin{table*}[t]
    \small
    \centering
    \begin{tabular}{c}
    \hline
        \textbf{Generated Responses} \\
        \hline
        \textbf{Query:} allan, i am really disappointed by your lack of effort on this project . \\
        \hline
        \textbf{VAE-AM:} it seems you are not aware of the actual work split . i have done more than anyone else. \\
        \textbf{DialogWAE:} ok , it's not that bad . but it's not that bad . i don't want to have any other questions about it . \\
        \hline
        \textbf{Query:} when is he coming ? \\
        \hline
        \textbf{VAE-AM:} next week . he is enjoying his month long vacation in miami . \\
        \textbf{DialogWAE:} sorry , he's not very serious . he's got a lot of money to do .  \\
        \hline
        \textbf{Query:} i ll try my best to make a good impression on the interviewer . \\
        \hline
        \textbf{VAE-AM:} do nt worry pal. only weak candidates should. just dress properly and try to make it alive there haha ... \\
        \textbf{DialogWAE:} that is ok , we ’ ll be able to get a raise in the next few days . we will have to pay you a lot of money . \\
        \hline
        \textbf{Query:} can you tell me where the pots and pans are ? \\
        \hline
        \textbf{VAE-AM:} absolutely not . i m not letting you cook today . you messed it real bad yesterday . \\
        \textbf{DialogWAE:} yes , of course .  \\
        \hline
        \textbf{Query:} what kind of personal information ? \\
        \hline
        \textbf{VAE-AM:} your name , height , sex ,  and contact . also tick this check box if you have faced this issue before . \\
        \textbf{DialogWAE:} not at the moment . but there is a lot of money in the city , but we have to pay a lot of money for the money . \\
        \hline
    \end{tabular}
    \caption{Generated responses in multi-turn setting}
    \label{tab:gen_multi}
\end{table*}

\begin{table*}[t]
    \small
    \centering
    \begin{tabular}{c|c|c|c|c|c|c|c|c|c}
    \hline
                \multirow{2}{*}{\textbf{Model}} 
                & \multicolumn{3}{|c|}{BLEU} & \multicolumn{5}{c|}{Diversity} & Fluency \\
                \cline{2-10}
                & \textbf{P} & \textbf{R} & \textbf{F} & \textbf{Intra-1} & \textbf{Intra-2} & \textbf{Inter-1} & \textbf{Inter-2} & \textbf{ASL} & \textbf{PPL}\\
                \hline
                \textbf{HRED*} & 0.232	& 0.232 & 0.232 & 0.94 & 0.97 & 0.09 & 0.09 & 10.1 & -
                \\
                \hline
                \textbf{CVAE*}	& 0.222 & 0.265 & 0.242 & 0.94 & 0.97 & 0.09 & 0.09 & 10.0 & - \\
                \hline
                \textbf{CVAE-CO*} & 0.244  & 0.259 & 0.251 & 0.94 & 0.97 & 0.09 & 0.09 & 11.2 & - \\
                \hline
                \textbf{VHCR*} & 0.266	& 0.289 & 0.277 & 0.85 & 0.97 & 0.42 & 0.74 & 16.9 & - \\
                \hline
                \textbf{DialogWAE} & 0.279	& 0.365 & 0.316 &	0.79 & 0.92 & 0.35 & 0.68 & 20.96 & 161.86                   \\
                \hline
                \textbf{VAE-AM (ours)} & \textbf{0.319} & \textbf{0.384} & \textbf{0.348}	&	0.93 & 0.95 & \textbf{0.49} & \textbf{0.93} & 15.1	& \textbf{127.39}   \\ \hline
    \end{tabular}
    \caption{DailyDialog Dataset: Multi-turn results. * denotes results taken from \cite{gu2018dialogwae} as is. Thus, the numbers correspond to evaluation on the non de-duplicated dataset and the PPL metric was also not reported.}
    \label{tab:dialog_multiturn_results}
\end{table*}

\bibliographystyle{coling}
\bibliography{coling2020}